\documentclass[conference]{IEEEtran}
\IEEEoverridecommandlockouts
\usepackage{cite}
\usepackage{amsmath,amssymb,amsfonts}
\usepackage{algorithmic}
\usepackage{graphicx}
\usepackage{textcomp}
\usepackage{xcolor}
\usepackage[ruled,linesnumbered]{algorithm2e}
\usepackage{booktabs}
\usepackage{multirow}
\usepackage{newtxtext}
\def\BibTeX{{\rm B\kern-.05em{\sc i\kern-.025em b}\kern-.08em
    T\kern-.1667em\lower.7ex\hbox{E}\kern-.125emX}}
\begin{document}

\title{GenCNER: A Generative Framework for Continual Named Entity Recognition
}

\author{\IEEEauthorblockN{1\textsuperscript{st} Yawen Yang}
\IEEEauthorblockA{\textit{School of Software} \\
\textit{Tsinghua University}\\
Beijing, China \\
yyw19@mails.tsinghua.edu.cn}
\and
\IEEEauthorblockN{2\textsuperscript{nd} Fukun Ma}
\IEEEauthorblockA{\textit{School of Software} \\
\textit{Tsinghua University}\\
Beijing, China \\
mfk22@mails.tsinghua.edu.cn}
\and
\IEEEauthorblockN{3\textsuperscript{rd} Shiao Meng}
\IEEEauthorblockA{\textit{School of Software} \\
\textit{Tsinghua University}\\
Beijing, China \\
msa21@mails.tsinghua.edu.cn \qquad}
\and
\IEEEauthorblockN{4\textsuperscript{th} Aiwei Liu}
\IEEEauthorblockA{\textit{School of Software} \\
\textit{Tsinghua University}\\
Beijing, China \\
liuaw20@mails.tsinghua.edu.cn}
\and
\IEEEauthorblockN{5\textsuperscript{th} Lijie Wen$^{*}$\thanks{$^{*}$Corresponding author.}}
\IEEEauthorblockA{\textit{School of Software} \\
\textit{Tsinghua University}\\
Beijing, China \\
wenlj@tsinghua.edu.cn}
}

\maketitle

\begin{abstract}
Traditional named entity recognition (NER) aims to identify text mentions into pre-defined entity types. Continual Named Entity Recognition (CNER) is introduced since entity categories are continuously increasing in various real-world scenarios. However, existing continual learning (CL) methods for NER face challenges of catastrophic forgetting and semantic shift of non-entity type. In this paper, we propose GenCNER, a simple but effective Generative framework for CNER to mitigate the above drawbacks. Specifically, we skillfully convert the CNER task into sustained entity triplet sequence generation problem and utilize a powerful pre-trained seq2seq model to solve it. Additionally, we design a type-specific confidence-based pseudo labeling strategy along with knowledge distillation (KD) to preserve learned knowledge and alleviate the impact of label noise at the triplet level. Experimental results on two benchmark datasets show that our framework outperforms previous state-of-the-art methods in multiple CNER settings, and achieves the smallest gap compared with non-CL results.
\end{abstract}


\section{Introduction}
Named Entity Recognition (NER) is one fundamental task in NLP fields due to its wide application in entity linking \cite{b1}, relation extraction \cite{b2} and knowledge graph \cite{b3}. Conventionally, NER involves recognizing text mentions into fixed pre-defined entity categories (e.g., ``Person'', ``Organization'') and the NER model only experiences one-time complete training process. While in real world, new entity types continue to emerge over time, which requires the model to perform incremental learning rather than training from scratch. As an example, voice assistants like Siri and Xiao Ai are frequently expected to extract new entity types for accurate understanding of user intents \cite{b4}, \cite{b5}. Thus the technique for Continual Named Entity Recognition (CNER) has attracted growing attention to learn new entity types incrementally without forgetting the old ones.
\begin{figure} 
  \centering
  \includegraphics[width=\linewidth]{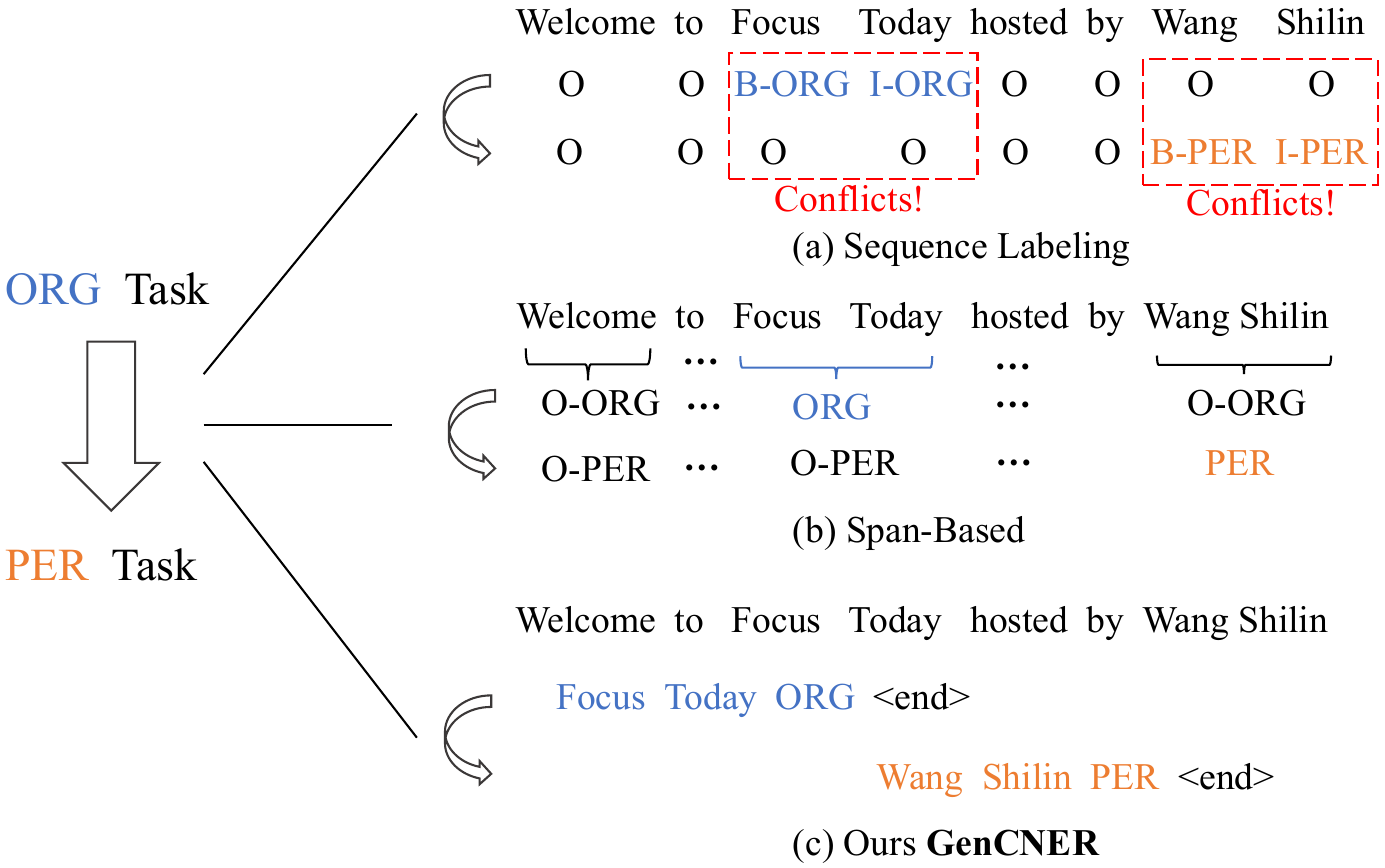}
  \caption{Changes of current ground-truth labels for sequence labeling, span-based and ours GenCNER method as CNER tasks increase. The red dashed box indicates conflicts between training targets, which leads to the semantic shift problem of the non-entity.}
  \label{fig1}
\end{figure}

Some efforts have been devoted to solve CNER tasks effectively. Existing works could be mainly divided into two strategies: sequence labeling and span-based models. The sequence labeling methods \cite{b4},\cite{b6},\cite{b7} assign each token one tag to indicate the entity boundary and type, and employ knowledge distillation (KD) to maintain old knowledge with entity type increasing. Nevertheless, such mechanism often encounters the semantic shift of non-entity type \cite{b5, b8}. In conventional NER labeling paradigm, tokens marked as non-entity do not belong to any entity type. While in CNER setting, the non-entity mark only indicates the corresponding tokens do not belong to current entity type(s). Consequently, the non-entity type will inevitably encompass the context token, previously learned entity types and future ones not yet encountered. As illustrated in Fig.~\ref{fig1}, besides true non-entity tokens, the non-entity mark ``O'' also covers future entity type ``PER'' in the current ``ORG'' task. The problem of semantic shift leads to inconsistent training objectives and exacerbates catastrophic forgetting among CL steps.

The span-based methods \cite{b9, b10} formulate CNER as a continuous binary classification task of the candidate entity spans and leverage knowledge distillation (KD) to retain memory. By changing from token-level labeling to span-level binary classification, the semantic shift of non-entity type could be avoided. As depicted in Fig.~\ref{fig1}, the non-entity mark regards entity span as non-(certain)-entity (``O-ORG''), which has no conflict with future entity type (``PER''). However, the basic framework proposed by Zhang and Chen \cite{b9} ignores label noise brought by previous teacher model. To address catastrophic forgetting and label noise problems, Chen and He \cite{b10} introduces Reinforcement Learning to span-based Knowledge Distillation for further improvement, but increases model complexity and computation costs.

The success of generative pre-trained language models (BART \cite{b11}, T5 \cite{b12}, etc.) has recently aroused interest in leveraging generative methods to solve information extraction tasks. Inspired by this, we propose \textbf{GenCNER}, a \textbf{Gen}erative framework with confidence-based pseudo labeling and knowledge distillation (KD) to reduce catastrophic forgetting and label noise. To be specific, we formulate continual NER task as sustained entity triplet label generation problem and employ the pre-trained seq2seq model BART with pointer network to generate target triplet sequence. In order to retain previously learned entity types and mitigate label noise, we first develop a confidence-based pseudo labeling strategy which removes pseudo entity triplets (predicted by teacher model) with lower confidence than type-specific thresholds, then empirically adopt knowledge distillation (KD) to transfer old knowledge to student model more accurately. 

Compared with sequence labeling methods for CNER, our generative framework GenCNER well avoids the problem of semantic shift since it has no need of non-entity marks in the generative background. As shown in Fig.~\ref{fig1}, the entity triplet is defined as ``(start token, end token, entity type)'' and the target sequence will continuously append triplets of new entity types as tasks increase. We can observe that the generative method updates learning targets for CNER task in a more natural and reasonable manner. Furthermore, GenCNER even supports both nested and discontinuous entity structures while other methods cannot or struggle to do so. Experimental results on OntoNotes and Few-NERD show that GenCNER achieves the highest F1 score at each CL step as well as the smallest performance gap to non-CL settings. The main contributions of this work can be summarized as follows:
\begin{enumerate}
\item[$\bullet$] We propose an effective generative framework to tackle CNER task, which innovatively transforms continual named entity recognition into consecutive entity triplet generation process. To the best of our knowledge, we are the first to introduce the generative mechanism to explorations of CNER tasks.
\item[$\bullet$] We develop a confidence-based pseudo labeling strategy with type-specific thresholds filtering to preserve high-quality entity triplets generated by old model for knowledge distillation (KD), weakening the influence of catastrophic forgetting and label noise.
\item[$\bullet$] We conduct adequate experiments on two public NER datasets in class incremental settings. Experimental results show that our proposed method outperforms strong baselines consistently on each dataset and achieves new state-of-the-art in different CL settings.
\end{enumerate}

\section{Related Work}
\subsection{Named Entity Recognition}
Named Entity Recognition (NER) aims to locate text mentions with specific meanings and classify them into predefined entity categories \cite{b13}. Recent works for various NER subtasks could be primarily divided into sequence labeling, span-based, hypergraph-based and seq2seq methods \cite{b14, b15}. The sequence labeling methods \cite{b16,b17} annotate each token with one tag indicating the entity boundary and type. Common tagging schemes include BIO and BIOES, where ``B'' represents the start of an entity, ``O'' stands for the context token not belonging to any entity. The span-based methods \cite{b18, b19} consider NER as a classification task of the candidate entity spans extracted from input sentence. Hypergraph-based methods \cite{b20} construct hypergraphs based on the entity structure and introduce graph neural network to learn semantic features. Most recently, the seq2seq framework which converts entity extraction into sequence generation has been proposed to solve multi-type NER subtasks uniformly \cite{b21}, achieving surprising performance.

\subsection{Continual Learning NER}
Continual Learning NER solves continuous tasks where training targets of new entity types emerge as a stream. Existing works treat Continual NER as the class-incremental problem \cite{b22} where the training data only contains labels of current entity type(s). Early explorations like AddNER and ExtendNER \cite{b6} employed the basic sequence labeling framework to tag tokens and knowledge distillation to prevent model from forgetting old entity types. To reduce the dependency of type co-occurrence during KD, Xia et al.\cite{b4} proposed L\&R, which provided ExtendNER with synthetic samples of old types for further distillation. Considering the semantic shift of non-entity type, Zheng et al.\cite{b7} designed a causal framework to retrieve the causality from both new entity types and Other-Class, Zhang et al.\cite{b5} developed a confidence-based pseudo-labeling for the non-entity type and Ma et al.\cite{b23} proposed to learn discriminative representations for entity types and ``O'' labels. On the other hand, SpanKL \cite{b9} and SKD-NER \cite{b10} introduced span-based classification to CNER, which avoided the problem of semantic shift and achieved satisfactory results.

\subsection{Seq2seq Framework}
The seq2seq model with encoder-decoder structure has been successfully adopted in various NLP tasks. It was first proposed by Cho et al.\cite{b24} in order to solve machine translation tasks. Bahdanau et al.\cite{b25} applied the attention mechanism to the decoder and obtained better translation performance. On this basis, the Pointer Network \cite{b26} and CopyNet \cite{b27} were proposed to get the token probability distributions of input sentences. With the increase of pre-trained language models, several attempts have been made to pre-train a seq2seq model, including MASS \cite{b28}, BART \cite{b11} and T5 \cite{b12}. In this paper, we focus on BART model due to its simplicity and promising performance in generative NLP tasks.

\section{Preliminary}
\subsection{Problem Formulation}
We follow previous works \cite{b6} to consider CNER as the class-incremental problem. We train the model on a sequence of tasks $T_1, T_2, ..., T_l$, where each task $T_k (1 \leq k \leq l)$ has its own training dataset $D_k$ annotated only for new entity types $E_k$. Note that entity types involved in different tasks are non-overlapping (i.e., $E_i \cap E_j = \emptyset$ when $i \neq j$). Meanwhile, sentences in each specific training dataset potentially also contain other entity types in previous or future steps, which are certainly not annotated in current task. At the first step, we train model $M_1$ 
on $D_1$ from scratch to identify entity types in $E_1$. At the $k$-th ($k > 1$) incremental step, our goal is to train a model $M_k$ based on the training set $D_k$ and previously learned model $M_{k-1}$ to recognize entities of all types seen so far, represented as $\cup_{i=1}^k E_i$.

\subsection{Learning Targets Construction}
In order to apply the generative framework to handle CNER task, we empirically represent each target entity as the triplet containing start token, end token and entity type. Given an input training sentence of $n$ tokens $X=[x_1, x_2, ..., x_n]$, the learning targets of different CNER tasks could be formulated as follows: At the first step, we expect to identify entities of types in $E_1$ and construct the target sequence $Y_1=[s_1, e_1, t_1, s_2, e_2, t_2, ..., s_g, e_g, t_g]$ to train the model $M_1$, where $g$ is the total number of entities, $s_i, e_i(1 \leq i \leq g)$ indicate the start and end token of the $i$-th entity respectively and $t_i \in E_1$ represents the entity type. At the $k$-th ($k > 1$) learning step with the previous model $M_{k-1}$ available, we first adopt $M_{k-1}$ to predict pseudo triplet sequence $Y_k^{p}$ including all the previously learned entity types $\cup_{i=1}^{k-1} E_i$, then connect the ground truth of triplet sequence $Y_k^{g}$ only containing $E_k$ types at the end of $Y_k^{p}$ to form the final target sequence $Y_k=[Y_k^{p}, Y_k^{g}]$ for model $M_k$ training.

\begin{figure*} 
  \centering
  \includegraphics[width=0.96\linewidth]{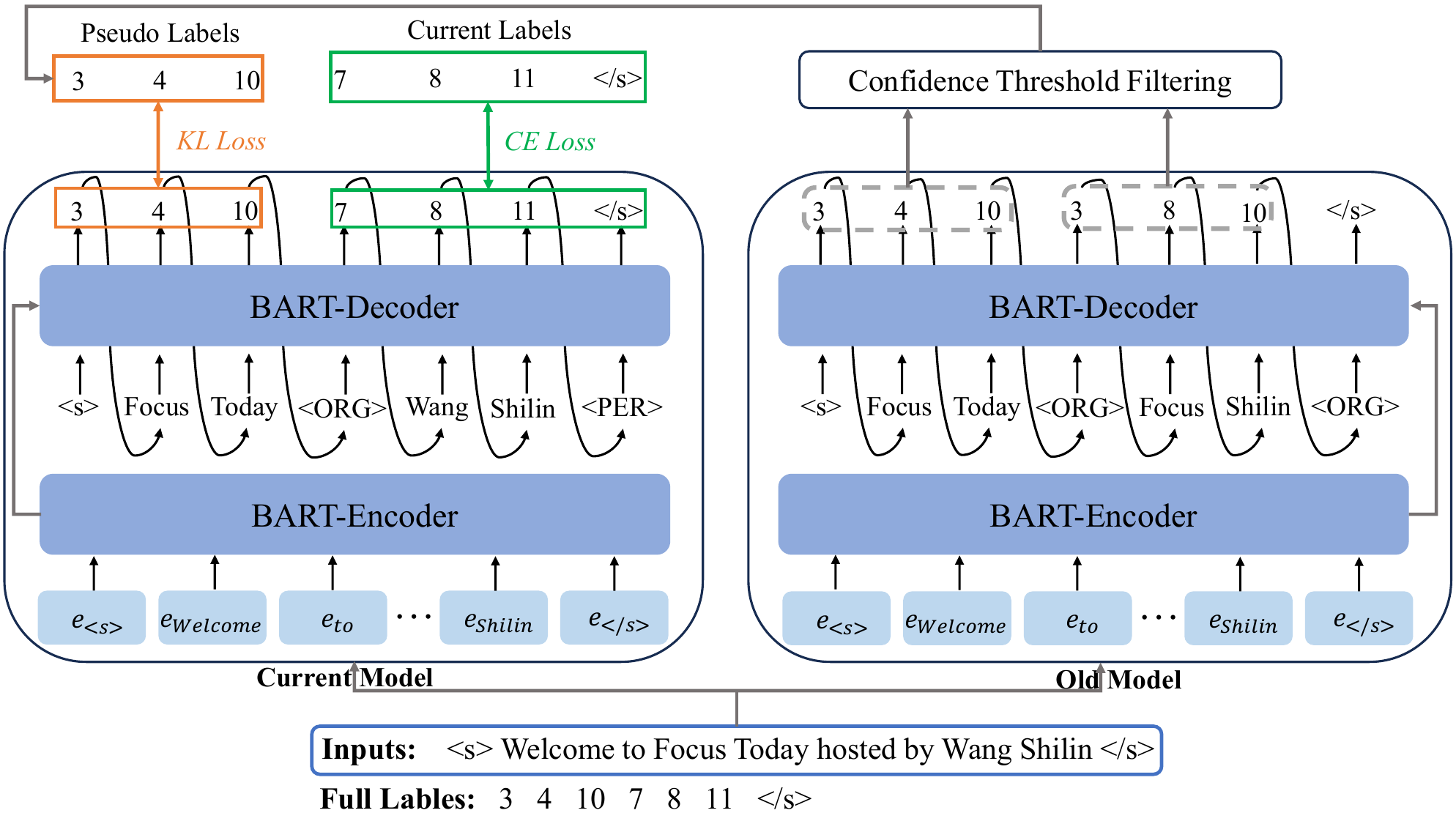}
  \caption{Overview of the proposed generative framework for Continual NER. Since the input sentence has 10 tokens, we conduct the 10 shift to entity type indexes. Thus index 0-9 indicates entity boundary tokens, and index larger than 9 represents different entity categories. $e_{<word>}$ denotes BART embeddings.}
  \label{fig2}
\end{figure*}

\section{Method}
The architecture of our framework is illustrated in Fig.~\ref{fig2}, consisting of two necessary components: Entity Triplet Sequence Generation (ETSG) and Confidence-based Pseudo Labeling (CBPL). In the first task of CNER, we train the model $M_1$ on $D_1$ with CE loss to make ETSG generate entity triplets of $E_1$ types. In other incremental tasks, we first use previous model to make a one-off prediction on training data, then prune the pseudo triplet sequence by confidence-based threshold filtering. We train the current model with KD loss and CE loss together to make ETSG capable of generating entity triplets of all seen types.

\subsection{Entity Triplet Sequence Generation}
As mentioned above, we need to predict entity triplet sequence in each incremental task of CNER. Considering previous works \cite{b21, b29} which effectively employ generative framework to solve the unified or multimodal NER tasks, we adopt the pre-trained seq2seq model BART with encoder-decoder structure to generate entity triplet sequence. Additionally, we introduce the idea of pointer network to BART decoder since the start and end tokens of each entity triplet both come from the input sentence. The BART encoder deals with the raw sentence $X=[x_1, x_2, ..., x_n]$ to get the contextual sentence-level representation $\mathbf{H}^e \in \mathbb R^{n \times d}$ as:
\begin{equation}
\setlength{\abovedisplayskip}{3pt}
\setlength{\belowdisplayskip}{3pt}
\mathbf{H}^e=\text{BartEncoder}(X),
\end{equation}
where $e$ represents the encoder output, $n$ is the sentence length and $d$ is the hidden dimension.

The BART decoder aims to generate boundary token or entity type index at each time step $P_t = P(y_t | H^e, Y_{<t})$. However, the previous triplet sequence $Y_{<t}$ cannot be put into the decoder module directly since it contains boundary token as well as entity type index. To tackle this problem, we continuously add new entity types as special tokens to the BART Tokenizer so that the sentence tokens and entity categories always share the same vocabulary. Then we develop an Index2Token mode to convert predicted indexes into practical tokens at each generative time step: 
\begin{equation}
\setlength{\abovedisplayskip}{3pt}
\setlength{\belowdisplayskip}{3pt}
\hat{y}_t = \begin{cases}
X_{y_t}, &if \  y_t \leq n \\
[\cup_{i=1}^k E_i]_{y_t-n}, &if \  y_t > n, 
\end{cases}
\end{equation}
where $[\cup_{i=1}^k E_i]$ represents all the entity types learned so far in the $k$-th incremental CNER task.

After converting indexes into tokens in this way, we can put the encoder output and previous triplet sequence into the decoder network and obtain the last hidden state, which formulates as:
\begin{equation}
\setlength{\abovedisplayskip}{3pt}
\setlength{\belowdisplayskip}{3pt}
\mathbf{h}^d_t=\text{BartDecoder}(\mathbf{H}^e;\hat{Y}_{<t}), 
\end{equation}
where $\hat{Y}_{<t}=[\hat{y}_1, \hat{y}_2, ..., \hat{y}_{t-1}]$, representing the generated index triplet sequence for time step $t$.

Then we apply the pointer mechanism to generate index probability distribution $P_t$ as:
\begin{align}
\mathbf{\hat{H}}^e &= \text{MLP}(\mathbf{H}^e), \\
{[\cup_{i=1}^k \mathbf{E}_i]}^d &= \text{BartTokenEmbed}(\cup_{i=1}^k \mathbf{E}_i), \\
\mathbf{L}_t &= [\mathbf{\hat{H}}^e \otimes \mathbf{h}^d_t;{[\cup_{i=1}^k \mathbf{E}_i]}^d \otimes \mathbf{h}^d_t], \\
\label{label-score} P_t &= \text{Softmax}(\mathbf{L}_t),
\end{align}
where the MLP module makes a linear transformation on the encoder output, BartTokenEmbed function means the token embeddings shared between encoder and decoder. Suppose the number of entity types in $\cup_{i=1}^k E_i$ is $ek$, then $\mathbf{\hat{H}}^e \in \mathbb R^{n \times d}$; ${[\cup_{i=1}^k \mathbf{E}_i]}^d \in \mathbb R^{ek \times d}$; $\mathbf{L}_t \in \mathbb R^{t \times (n+ek)}$; $\otimes$ means the dot product of multi-dimension vectors and $[\cdot;\cdot]$ concats the two vectors in the first dimension.

\subsection{Confidence-based Knowledge Distillation}
GenCNER uses knowledge distillation to avoid forgetting old types of entities while learning the new ones. At the $k$-th incremental task, we make a one-off prediction with the previously learned model $M_{k-1}$ on the current training dataset for all the seen types $\cup_{i=1}^{k-1} E_i$. Next, we connect the pseudo triplet sequence generated by model $M_{k-1}$ with the annotated triplet sequence of current entity type(s) $E_k$, to construct target triplet sequence for model $M_{k}$ optimization. During KD process, the previous model $M_{k-1}$ plays the role of teacher to distill old knowledge into the current student $M_{k}$.

It is worth noting that the pseudo triplet sequence possibly brings label noise which interferes the training of current model. In order to reduce model noise and improve distillation efficiency, we empirically design a confidence-based pseudo labeling mechanism for the filtering of poor predictions. Specifically, towards each training sentence in $D_k$, we first compute the probability of each token predicted by $M_{k-1}$. Since it is difficult to evaluate the quality of one single token, we split the probability sequence every three generative steps, corresponding to each predicted entity triplet. Then we obtain the minimum value of each probability triplet to stand for the generation quality of different entity triplets. Since different entity types have diverse learning difficulties, we define the confidence threshold in a type-specific way referring to previous semi-supervised works \cite{b30, b31}.
\begin{equation}
\delta^{(c)} = \text{min}(\delta, \text{median}\{\text{min}(p(s), p(e), p(c)) | (s,e,c) \in S^{k-1}\}),
\label{thre_equa}
\end{equation}
where $\delta$ is the type-specific hyper-parameter, $s$(or $e$) denotes the start(or end) token of a predicted entity, $c \in \cup_{i=1}^{k-1} E_i$ represents one specific entity type and $S^{k-1}$ denotes the whole entity triplet set predicted by $M_{k-1}$. Such a threshold ensures that we keep at least 50\% pseudo triplets of each entity type to train a new model. 

For noise minimization, entity triplets with token probability lower than the type-specific thresholds are removed from predicted entity triplet sequence, forming the new pseudo label sequence as follows: 
\begin{equation}
Y_k^{pru} = \{(s_i, e_i, t_i)\ |\ \text{min}(p(s_i), p(e_i), p(t_i)) \geq \delta^{(t_i)}\},
\end{equation}
where $(s_i, e_i, t_i) \in Y_k^p$ denotes the $i$-th entity triplet generated by the old model $M_{k-1}$.

The pruned pseudo triplet sequence $Y_k^{pru}$ is used to compute the KL divergence loss of soft labels with the current model $M_{k}$, which formulates as:
\begin{equation}
\setlength{\abovedisplayskip}{3pt}
\setlength{\belowdisplayskip}{3pt}
\mathcal{L}_{KD} = \frac{1}{L}\sum_{t=1}^{L}P(y_t^{pru})(\log P(y_t^{pru})-\log f_\theta(X, \hat{Y}_{<t})),
\end{equation}
where $L$ is the length of pruned pseudo triplet sequence, $y_t^{pru} \in Y_k^{pru}$ stands for the boundary token or entity type predicted by old model $M_{k-1}$ at the $t$-th generative step, $P(y_t^{pru})$ means the corresponding probability distribution of $y_t^{pru}$, $f_\theta$ represents the current student model $M_{k}$.

\subsection{Model Training}
Considering that we expect the model to learn new entity types by extending entity triplet sequence, the cross entropy loss is computed as:
\begin{equation}
\setlength{\abovedisplayskip}{3pt}
\setlength{\belowdisplayskip}{3pt}
\mathcal{L}_{CE} = -\frac{1}{G}\sum_{t=L+1}^{L+G}P_{one-hot}(y_t^g)\log f_\theta(X, \hat{Y}_{<t}),
\end{equation}
where $G$ represents the length of current ground truth of triplet sequence, $y_t^g \in Y_k^g$ means the $(t-L)$-th value of current annotated triplet sequence and $f_\theta$ represents the current model $M_{k}$, as mentioned above.

Following previous continue learning methods for NER \cite{b6, b9}, the final loss in the $k$-th incremental task $(k>1)$ is the weighted sum as:
\begin{equation}
\setlength{\abovedisplayskip}{3pt}
\setlength{\belowdisplayskip}{3pt}
\mathcal{L} = \alpha \mathcal{L}_{KD} + \beta \mathcal{L}_{CE},
\end{equation}
where $\alpha$ and $\beta$ are both the weights of different losses to balance the learning of past and present.

\section{Experiments}
We conduct extensive experiments on two public datasets to prove the effectiveness of our proposed model in recognizing entities for growing CNER tasks. Then the detailed analyses are given to show the strengths of each component.

\subsection{Datasets}
For the evaluation of proposed generative framework, we conduct main experiments on two public NER datasets: OntoNotes \cite{b32} and Few-NERD \cite{b33}. To imitate the class-incremental setting, we follow previous works \cite{b4, b6} to split the above standard NER corpora into separate datasets used in a series of CL tasks. Compared with OntoNotes which only annotates one single entity type within each task, the fine-grained dataset Few-NERD allows learning multiple types per task. Some basic information of them is described below. \textbf{OntoNotes-5.0 English} is one flat NER dataset in the field of news, containing 18 entity categories. We follow recent works \cite{b9} to select 6 types to ensure sufficient training samples, including Organization, Person, GPE, Date, Cardinal and NORP. Each entity type is annotated only in the corresponding CL task. \textbf{Few-NERD} is first proposed for few-shot NER research, which consists of 8 coarse-grained and 66 fine-grained entity types as presented in Table \ref{fine_grained}. For simplicity, we employ the supervised full version of Few-NERD and construct a sequence of CL tasks via each coarse-grained type. Consequently, each task can be seen as a semantic domain with some relative fine-grained entity types, more similar to the complicated applications in the real world.

\begin{table}
\caption{\label{fine_grained}
The fine-grained entity types of Few-NERD.
}
\centering
\resizebox{\linewidth}{!}{
\begin{tabular}{l|l}
\toprule
Coarse-grained category & Fine-grained types \\
\midrule
\multirow{2}*{location (LOC)} & GPE, park, island, mountain, bodiesofwater, \\ 
~ & road/railway/highway/transit, other \\
\midrule
\multirow{2}*{person (PER)} & actor, athlete, scholar, director, soldier, \\ 
~ & politician, artist/author, other \\
\midrule
\multirow{3}*{organization (ORG)} & company, religion, education, sportsteam, \\
~ & sportsleague, politicalparty, showorganization, \\
~ & media/newspaper, government, other \\
\midrule
\multirow{3}*{other (OTH)} & god, law, astronomything, award, biologything, \\
~ & chemicalthing, currency, disease, language, \\
~ & educationaldegree, livingthing, medical \\
\midrule
\multirow{2}*{product (PROD)} & airplane, car, food, game, ship, \\
~ & software, train, weapon, other \\
\midrule
\multirow{2}*{build (BUID)} & airport, hospital, hotel, restaurant, \\
~ & library, sportsfacility, theater, other \\
\midrule
\multirow{2}*{art (ART)} & broadcastprogram, film, music, \\
~ & painting, writtenart, other \\
\midrule
\multirow{2}*{event (EVET)} & attack/battle/war/militaryconflict, disaster, \\
~ & election, protest, sports, other \\
\bottomrule
\end{tabular}
}
\end{table}

\subsection{CNER Settings}
Before experimental evaluation, we need to split the original dataset reasonably to construct a series of CL dataset. Nevertheless, we discover the obvious diversity of different split methods in previous works, thus making their results difficult to compare. Following Zhang and Chen \cite{b9}, the diversity of CL datasets includes two aspects: \textbf{1)} To divide the initial train/dev dataset into a series of train/dev dataset for each CL task, Monaikul et al. \cite{b6} separates original samples randomly into disjoint datasets, while Xia et al. \cite{b4} filters samples having entity types designated to be learned in that corresponding task. The above division patterns are defined as \emph{Split} and \emph{Filter}, respectively. \textbf{2)} When constructing the test dataset evaluated in each CL task, Monaikul et al. \cite{b6} directly adopt the full version of initial test set, while Xia et al. \cite{b4} still filters test samples having one or more entity categories learned so far. The different splits of test data are represented as \emph{All} and \emph{Filter}, similarly. 

Armed with these observations, there exist four combinations of CNER training and testing: \emph{Split-All}, \emph{Split-Filter}, \emph{Filter-All}, \emph{Filter-Filter}. We select \emph{\textbf{Split-All}} and \emph{\textbf{Filter-Filter}} for comprehensive evaluation of proposed framework and reliable comparisons with previous works.

\subsection{Baselines and Evaluation metrics}
We compare GenCNER with typical baselines utilizing different recognizing mechanisms. \textbf{1) Sequence labeling}: \textbf{AddNER} and \textbf{ExtendNER} \cite{b6} are the early state-of-the-art works to apply sequence labeling and knowledge distillation to CNER task. \textbf{L\&R} \cite{b4} enhances ExtendNER with an extra reviewing stage of synthetic augmented samples. We can not compare with CFNER \cite{b7}, CPFD \cite{b5} and O\_CILNER \cite{b23} since they employ data sampling operations and different CL settings. \textbf{2) Span-based}: \textbf{SpanKL} \cite{b9} is the first to introduce span-based classification to achieve coherent optimization in CNER. \textbf{SKD-NER} \cite{b10} employs reinforcement learning strategies during span-based KD process to optimize the soft labeling and distillation loss.

For evaluation metrics, we employ the Macro F1 score of all seen types up to each learning step, and report the average results of tests on different learning orders predefined in Table~\ref{learn_order}. Since the fine-grained types in Few-NERD are not balanced, we first compute the Micro F1 of each coarse-grained type then fairly report the Macro F1 of all seen coarse-grained types. Additionally, we report the gap between non-CL and CL results in each incremental task. The non-CL setting adds all the previous training datasets to the current one and provides full annotations of entity types learned so far, which serves as the upper bound of F1 performance at each step.

\begin{table}
\caption{\label{learn_order}
Different learning orders of CL tasks on two datasets. 
}
\centering
\resizebox{\linewidth}{!}{ 
\begin{tabular}{rcccccccc}
\toprule
\multicolumn{9}{c}{6 OntoNotes Permutations ($\rightarrow$)} \\
\midrule
1:  & ORG & PER & GPE & DATE & CARD & NORP &  &  \\
\midrule
2:  & DATE & NORP & PER & CARD & ORG & GPE &  &  \\
\midrule
3:  & GPE & CARD & ORG & NORP & DATE & PER &  &  \\
\midrule
4:  & NORP & ORG & DATE & PER & GPE & CARD &  &  \\
\midrule
5:  & CARD & GPE & NORP & ORG & PER & DATE &  &  \\
\midrule
6:  & PER & DATE & CARD & GPE & NORP & ORG &  &  \\
\midrule
\multicolumn{9}{c}{4 Few-NERD Permutations ($\rightarrow$)} \\
\midrule
1:  & LOC & PER & ORG & OTH & PROD & BUID & ART & EVET \\
\midrule
2:  & ORG & PROD & ART & EVET & OTH & PER & LOC & BUID \\
\midrule
3:  & PROD & EVET & OTH & PER & ART & LOC & BUID & ORG \\ 
\midrule
4:  & BUID & OTH & PROD & PER & ORG & LOC & ART & EVET \\
\bottomrule
\end{tabular}
}
\end{table}

\subsection{Implementation Details}
During data preparation, we continuously add entity types as special tokens to the BART Tokenizer with CL tasks increasing. Meanwhile, we apply the $n$ shift ($n$ is the length of input sentences after padding) to each entity type index, which helps to identify whether the predicted index is a boundary token or an entity type.

Towards model components, we apply the pre-trained seq2seq model BART-large to continuously generate entity triplet sequence. We only use BART embeddings with the dimension of 1024. During confidence-based pseudo labeling, we remove entity triplets predicted by old model with lower confidence than type-specific thresholds defined in formula \ref{thre_equa}. Considering the diverse learning difficulties of different entity types and the trade-off between quantity and quality of training samples, we select the threshold hyperparameter $\delta$ according to Table~\ref{threshold_setting}. To effectively balance KD and CE losses, the corresponding weights are set to 1($\alpha$) and 0.5($\beta$). 

In model training, we choose the AdamW optimizer with the learning rate 5e-5 and set warmup ratio to 0.1 for both datasets. The epoch number of each CL task is 10 and the batch size is set to 8. We perform each experiment on a single GeForce RTX 3090 GPU.

\begin{table}
\caption{\label{threshold_setting}
Type-specific threshold settings of two datasets. We test a range of thresholds on the validation set and select the optimal one for each entity type.
}
\centering
\resizebox{0.88\linewidth}{!}{
\begin{tabular}{c|c|c}
\toprule
Dataset & Involved Entity Type & Threshold $\delta$ \\
\midrule
\multirow{6}*{OntoNotes 5.0} & ORG & 0.78 \\
~ & PER & 0.84 \\
~ & GPE & 0.74 \\
~ & DATE & 0.71 \\
~ & CARD & 0.63 \\
~ & NORP & 0.77 \\
\midrule
\multirow{8}*{Few-NERD} & LOC & 0.60 \\
~ & PER & 0.58 \\
~ & ORG & 0.54 \\
~ & OTH & 0.54 \\
~ & PROD & 0.49 \\
~ & BUID & 0.45 \\
~ & ART & 0.46 \\
~ & EVET & 0.51 \\
\bottomrule
\end{tabular}
}
\end{table}

\begin{table*}
\caption{\label{main_result_onto}
Macro F1 of different models at each CL step in \emph{Split-All} and \emph{Filter-Filter} settings on OntoNotes. $\Delta=\text{CL}-\text{non-CL}$. The highest F1 score at final learning step is bolded along with its gap. All methods adopt the large version of BERT or BART as encoder, ensuring their parameter sizes are at the same level.
}
\centering
\resizebox{\linewidth}{!}{ 
\begin{tabular}{c|cccccc|cccccc}
\toprule
\multirow{2}*{Method} & \multicolumn{6}{c}{Train(\textbf{Split})-Test(\textbf{All})} & \multicolumn{6}{c}{Train(\textbf{Filter})-Test(\textbf{Filter})} \\
\cmidrule(r){2-7}
\cmidrule(r){8-13}
~ & Step1 & Step2 & Step3 & Step4 & Step5 & Step6 & Step1 & Step2 & Step3 & Step4 & Step5 & Step6 \\
\midrule
\color{red}{non-CL} & 82.52 & 86.42 & 87.32 & 88.84 & 89.62 & 89.27 & 90.78 & 91.54 & 90.76 & 90.60 & 90.50 & 90.48 \\
AddNER \cite{b6} & 82.52 & 83.90 & 84.66 & 85.02 & 85.48 & 85.03 & 90.78 & 89.82 & 88.92 & 87.20 & 86.16 & 85.82 \\
$\Delta$ & -0.00 & -2.52 & -2.66 & -3.82 & -4.14 & -4.24 & -0.00 & -1.72 & -1.84 & -2.40 & -3.34 & -4.66 \\
\midrule
\color{red}{non-CL} & 82.79 & 86.49 & 87.70 & 88.46 & 89.02 & 89.19 & 90.62 & 91.70 & 91.02 & 90.79 & 90.92 & 90.10 \\
ExtendNER \cite{b6} & 82.79 & 83.54 & 84.48 & 84.67 & 85.12 & 84.96 & 90.62 & 88.92 & 87.55 & 86.30 & 84.77 & 81.37 \\
$\Delta$ & -0.00 & -2.95 & -3.22 & -3.79 & -3.90 & -4.23 & -0.00 & -2.78 & -3.47 & -4.49 & -6.15 & -8.73 \\
\midrule
\color{red}{non-CL} & - & - & - & - & - & - & 92.06 & 91.16 & 90.50 & 89.69 & 89.57 & 89.30 \\
L\&R \cite{b4} & - & - & - & - & - & - & 92.06 & 88.09 & 85.69 & 83.79 & 83.38 & 83.02 \\
$\Delta$ & - & - & - & - & - & - & -0.00 & -3.07 & -4.81 & -5.90 & -6.19 & -6.28 \\
\midrule
SKD-NER \cite{b10} & 87.33 & 91.47 & 92.36 & 90.76 & 88.54 & 88.17 & - & - & - & - & - & - \\
\midrule
\color{red}{non-CL} & 85.60 & 88.16 & 88.64 & 89.39 & 89.69 & 89.74 & 92.37 & 92.65 & 92.78 & 92.06 & 92.10 & 91.90 \\
SpanKL \cite{b9} & 85.60 & 87.92 & 88.22 & 88.76 & 89.02 & 88.98 & 92.37 & 90.81 & 90.38 & 89.50 & 89.18 & 88.07 \\
$\Delta$ & -0.00 & -0.24 & -0.42 & -0.63 & -0.67 & -0.76 & -0.00 & -1.84 & -2.40 & -2.56 & -2.92 & -3.83 \\
\midrule
\color{red}{non-CL} & 86.37 & 89.26 & 89.41 & 89.93 & 90.40 & 90.57 & 92.79 & 93.48 & 93.67 & 92.96 & 92.91 & 92.23 \\
\textbf{GenCNER(ours)} & 86.37 & 88.84 & 89.06 & 89.42 & 89.71 & \textbf{89.95} & 92.79 & 91.86 & 91.39 & 90.13 & 89.73 & \textbf{88.76} \\
$\Delta$ & -0.00 & -0.42 & -0.35 & -0.51 & -0.69 & \textbf{-0.62} & -0.00 & -1.62 & -2.28 & -2.83 & -3.18 & \textbf{-3.47} \\
\bottomrule
\end{tabular}
}
\end{table*}

\subsection{Main Results}
We evaluate GenCNER performance on OntoNotes in two CNER settings: \emph{Split-All} and \emph{Filter-Filter}. The former randomly splits the original training dataset into disjoint subsets and adopts the full version of initial test set, which is the most common setup for CNER. The latter selects samples having current entity types from initial training set and filters samples with types seen so far in original test set, lacking training samples without any entity.

\begin{table}
\caption{\label{main_result_fewn}
Macro F1 on Few-NERD in the \emph{Split-All} CL setting. Each step involves one coarse-grained entity type containing some fine-grained ones.
}
\centering
\resizebox{\linewidth}{!}{ 
\begin{tabular}{c|cccccccc}
\toprule
\multirow{2}*{Method} & \multicolumn{8}{c}{Train(\textbf{Split})-Test(\textbf{All})} \\
\cmidrule(r){2-9}
~ & Step1 & Step2 & Step3 & Step4 & Step5 & Step6 & Step7 & Step8 \\
\midrule
\color{red}{non-CL} & 64.01 & 62.25 & 61.88 & 61.07 & 61.28 & 62.43 & 63.83 & 63.65 \\
AddNER \cite{b6} & 64.01 & 61.32 & 60.54 & 59.43 & 58.74 & 59.32 & 60.41 & 59.32 \\
$\Delta$ & -0.00 & -0.93 & -1.34 & -1.64 & -2.54 & -3.11 & -3.42 & -4.33 \\
\midrule
\color{red}{non-CL} & 64.06 & 62.08 & 61.76 & 60.94 & 61.29 & 62.02 & 63.62 & 63.32 \\
ExtendNER \cite{b6} & 64.06 & 59.02 & 57.05 & 55.72 & 55.46 & 55.96 & 56.85 & 56.16 \\
$\Delta$ & -0.00 & -3.06 & -4.71 & -5.22 & -5.83 & -6.06 & -6.77 & -7.16 \\
\midrule
L\&R \cite{b4} & 68.13 & 66.72 & 64.51 & 63.44 & 60.97 & 61.23 & 60.88 & 60.32 \\
\midrule
\color{red}{non-CL} & 67.81 & 65.22 & 64.97 & 64.18 & 64.22 & 64.94 & 66.10 & 65.76 \\
SpanKL \cite{b9} & 67.81 & 64.16 & 63.62 & 62.31 & 61.67 & 62.17 & 63.24 & 62.15 \\
$\Delta$ & -0.00 & -1.06 & -1.35 & -1.86 & -2.55 & -2.77 & -2.86 & -3.61 \\
\midrule
\color{red}{non-CL} & 68.27 & 65.73 & 65.36 & 64.63 & 64.92 & 65.48 & 66.52 & 66.31 \\
\textbf{GenCNER(ours)} & 68.27 & 64.75 & 64.10 & 63.02 & 62.28 & 62.95 & 63.74 & \textbf{62.96} \\
$\Delta$ & -0.00 & -0.98 & -1.26 & -1.61 & -2.64 & -2.53 & -2.78 & \textbf{-3.35} \\
\bottomrule
\end{tabular}
}
\end{table}

\begin{table*}
\caption{\label{ablation_onto}
The ablation study on OntoNotes in two CL settings. ``\emph{CTF}'' represents Confidence Threshold Filtering. ``w/o $\mathcal{L}_{KD}$'' denotes replacing KL divergence with cross entropy loss At the KD stage.
}
\centering
\resizebox{0.95\linewidth}{!}{ 
\begin{tabular}{l|cccccc|cccccc}
\toprule
\multirow{2}*{Method} & \multicolumn{6}{c}{Train(\textbf{Split})-Test(\textbf{All})} & \multicolumn{6}{c}{Train(\textbf{Filter})-Test(\textbf{Filter})} \\
\cmidrule(r){2-7}
\cmidrule(r){8-13}
~ & Step1 & Step2 & Step3 & Step4 & Step5 & Step6 & Step1 & Step2 & Step3 & Step4 & Step5 & Step6 \\
\midrule
\textbf{GenCNER} & 86.37 & 88.84 & 89.06 & 89.42 & 89.71 & \textbf{89.95} & 92.79 & 91.86 & 91.39 & 90.13 & 89.73 & \textbf{88.76} \\
w/o \emph{CTF}  & 86.37 & 88.56 & 88.67 & 89.10 & 89.38 & 89.43 & 92.79 & 91.14 & 90.43 & 89.07 & 88.68 & 87.41 \\
w/o $\mathcal{L}_{KD}$ & 86.37 & 88.65 & 88.81 & 89.26 & 89.32 & 89.58 & 92.79 & 91.46 & 90.72 & 89.62 & 89.23 & 88.29 \\
\bottomrule
\end{tabular}
}
\end{table*}

As shown in Table \ref{main_result_onto}, GenCNER significantly outperforms other baselines in both CNER settings. It not only achieves the highest Macro F1 scores at each step, but also has the smallest gap to the non-CL performance. Compared with previous SOTA method SpanKL, GenCNER achieves +0.97\%, +0.69\% F1 higher, and narrows the performance gap by the absolute value of 0.14, 0.36 at the final learning step in \emph{Split-All} and \emph{Filter-Filter} settings, respectively. Experimental results on OntoNotes demonstrate the effectiveness of GenCNER to recognize entities of growing types. We also discover that different methods perform similarly in non-CL setup but quite diversely for CNER, revealing they may be strong enough to solve traditional NER but have varying degrees of difficulty in handling a series of CL tasks.

Following recent works \cite{b4, b9}, we test GenCNER on a more complicated dataset Few-NERD under typical \emph{Split-All} CL setting and display the results in Table \ref{main_result_fewn}. Each step represents one coarse-grained entity type, which contains several fine-grained ones and well imitates real-life scenarios. From Table \ref{main_result_fewn}, we observe that although facing the challenge of learning multiple fine-grained types per task, GenCNER still performs better than other baselines. To be specific, GenCNER improves the F1 score by 0.81\% at the final step and reduces the gap from -3.61 to -3.35.

We briefly analyze the core factors that lead to performance improvement. As we convert CNER into sequence generation of rising entity triplets, the problem of semantic shift no longer exists, thus achieving coherent optimization. During KD stage, we abandon pseudo triplets with lower confidence in a type-specific way, which reduces the label noise and prevents subsequent error accumulation.

\subsection{Further Analysis}
\textbf{Ablation Study} To explore the effectiveness of confidence-based filtering strategy and KL divergence loss of soft pseudo labels, we conduct necessary ablation experiments on OntoNotes. As shown in Table \ref{ablation_onto}, when we remove confidence threshold filtering from GenCNER, the F1 score decreases by 0.52\%, 1.35\% at the final step. Meanwhile, we find the \emph{Filter-Filter} setting suffers more obvious performance decline. That may be because \emph{Filter-Filter} constructs training sets by filtering samples with specific entity types, causing intersection between different sets. Thus there exist more samples of type co-occurence in \emph{Filter-Filter} CL setting, potentially bringing more noise during old model prediction. With regard to pseudo label format and optimized objectives, the F1 score reveals a slight decrease if we replace KL divergence with cross entropy loss, indicating the strength of soft label learning, as expected. \\

\begin{figure*} 
  \centering
  \includegraphics[width=\linewidth]{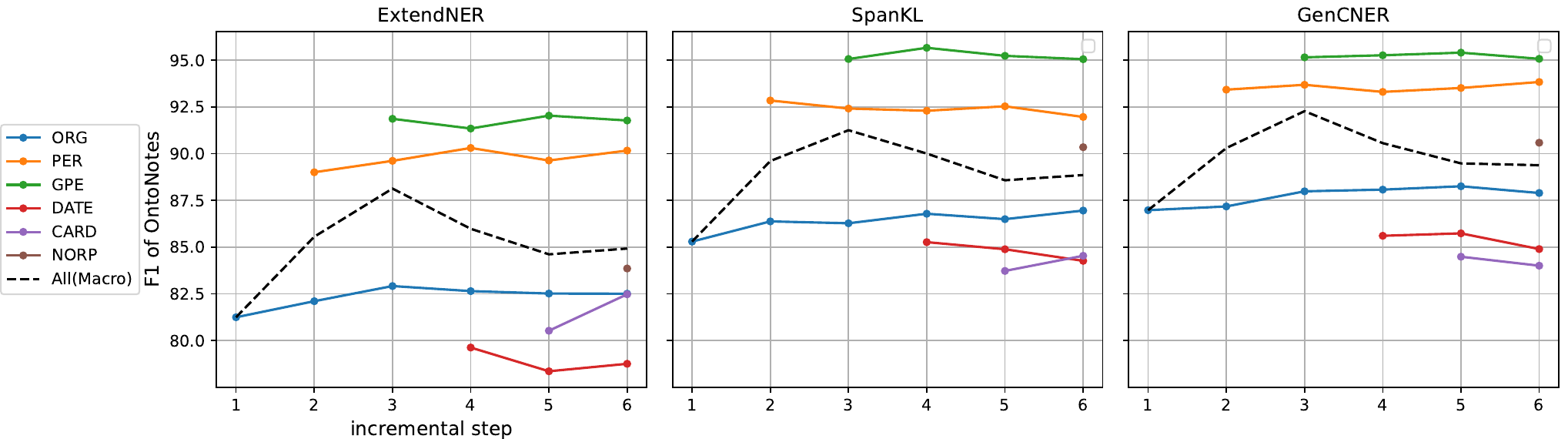}
  \caption{F1 curves of involved OntoNotes entity type(s) at each step with a certain learning order in \emph{Split-All} setup.}
  \label{fig3}
\end{figure*}

\textbf{Type-Level Comparison} We further study the instant performance of involved entity types at each CL step. Following a certain learning order, We plot their F1 curves on OntoNotes in \emph{Split-All} setting and add an extra dashed curve to denote the overall Macro F1 up to per step. As illustrated in Fig.~\ref{fig3}, GenCNER outperforms other baselines in the following aspects: 1) The starting point of each curve in GenCNER is always superior to that in other methods, indicating the stronger power of GenCNER in acquiring new type knowledge. 2) The trend of most F1 curves in GenCNER is obviously smoother with CL tasks growing, which reveals it effectively prevents the new model from forgetting already learned entity categories. \\

\textbf{Effect of triplet compositions} During entity triplet sequence generation, we utilize the classic autoregressive framework, where current decoder input depends on the output token from last time step. Considering the influence of conditional probability in seq2seq models, we evaluate three compositions of entity triplet on Ontonotes in \emph{Split-All}: TSE (type, start, end), STE and SET. From Table \ref{triplet_com}, we find that the composition of triplet has a certain impact on model performance and SET defined as (start, end, type) achieves the best F1, indicating the recognition of boundary tokens promotes the classification of entity types.\\

\begin{table}
\caption{\label{triplet_com}
Macro F1 of different entity triplet compositions on OntoNotes in \emph{Split-All} setting.
}
\centering
\resizebox{0.95\linewidth}{!}{ 
\begin{tabular}{r|cccccc}
\toprule
\multirow{2}*{Method} & \multicolumn{6}{c}{Train(\textbf{Split})-Test(\textbf{All})} \\
\cmidrule(r){2-7}
~ & Step1 & Step2 & Step3 & Step4 & Step5 & Step6 \\
\midrule
TSE & 86.05 & 88.46 & 88.77 & 88.89 & 89.32 & 89.63 \\
STE & 86.16 & 88.53 & 88.70 & 89.22 & 89.48 & 89.79 \\
\textbf{SET} & 86.37 & 88.84 & 89.06 & 89.42 & 89.71 & \textbf{89.95} \\
\bottomrule
\end{tabular}
}
\end{table}

\begin{figure} 
  \centering
  \includegraphics[width=0.96\linewidth]{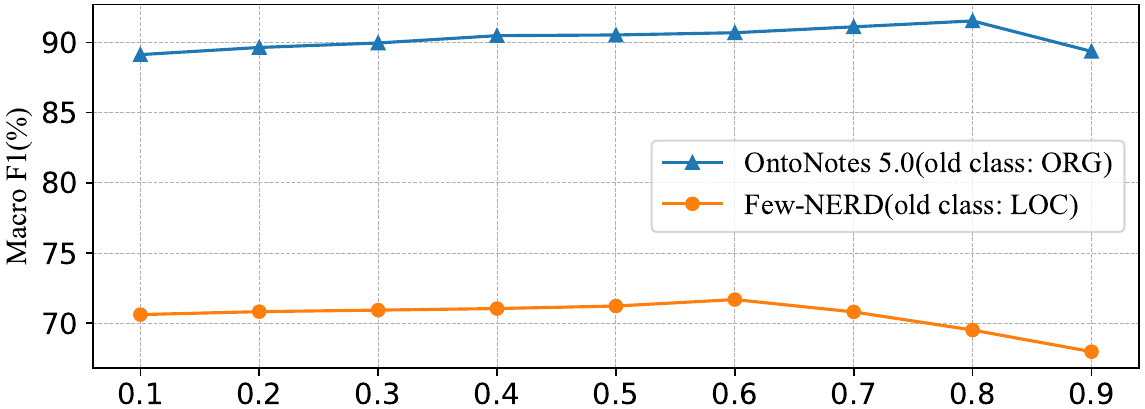}
  \caption{Effect of type-specific threshold $\delta$ selection at the second CL step.}
  \label{fig4}
\end{figure}

\textbf{Analysis of Threshold Selection} We finally study the selection of type-specific threshold $\delta$ on two datasets at the second CL step in \emph{Split-All} setting. Fig.~\ref{fig4} reveals that F1 score first increases then decreases as the threshold increases progressively. Such trend indicates that reasonable threshold settings can help reduce label noise, while excessive thresholds may lead to insufficient training samples. Additionally, it is necessary to set confidence thresholds in a type-specific way since different entity categories have diverse optimal thresholds for knowledge distillation.

\section{Conclusion}
In this paper, we propose a novel generative framework for continual NER. By formulating CNER into sequence generation of increasing entity triplets, GenCNER averts the semantic shift of non-entity and solves a series of CL tasks excellently. We perform type-specific confidence threshold filtering to select high-quality pseudo triplets generated by old model, which effectively reduces the potential noise and enhances knowledge distillation. Experiments on two NER datasets show GenCNER achieves new SOTA performance over strong baselines in multiple CL settings.

Meanwhile, our framework also has some limitations. Compared with other methods, GenCNER costs more training and inference time since it adopts the autoregressive structure to generate triplet sequences in each CL task. In addition, the knowledge distillation greatly depends on the co-occurrence of entity types between different tasks. The model performance will significantly decrease if the current task contains fewer previously learned entity types. We plan to explore the above issues for future work.

\section*{Acknowledgment}
This work is supported by the National Key Research and Development Program of China (No.2024YFB3309702), the National Nature Science Foundation of China (No.62021002), Tsinghua BNRist and Beijing Key Laboratory of Industrial Big Data System and Application.


\vspace{12pt}

\end{document}